\documentclass{Interspeech}

\interspeechcameraready 

\title{Speech-to-Text Translation with Phoneme-Augmented CoT:\\Enhancing Cross-Lingual Transfer in Low-Resource Scenarios}

\author[affiliation={1,2}]{Gerard I.}{Gállego$^{\dagger}$}
\author[affiliation={1}]{Oriol}{Pareras$^{\dagger}$}
\author[affiliation={1}]{Martí}{Cortada Garcia}
\author[affiliation={1}]{Lucas}{Takanori}
\author[affiliation={1,2}]{Javier}{Hernando}

\affiliation{}{Barcelona Supercomputing Center}{Spain}
\affiliation{}{Universitat Politècnica de Catalunya}{Spain}
\email{gerard.gallego@bsc.es, oriol.pareras@bsc.es}

\usepackage{comment}
\usepackage{siunitx}
\usepackage{multirow}
\usepackage{float}
\usepackage[most]{tcolorbox}
\usepackage{lipsum}
\usepackage{nth}

\DeclareSIUnit{\million}{\text{million}}

\newcommand{\ourmodelname}{\textsc{Salamandra-ST}}

\newcommand{\ourmodel}{\ourmodelname\xspace}
\newcommand{\ourmodeldirect}{\ourmodelname\textsubscript{\textsc{direct}}\xspace}
\newcommand{\ourmodelcot}{\ourmodelname\textsubscript{\textsc{CoT}}\xspace}
\newcommand{\ourmodelphcot}{\ourmodelname\textsubscript{\textsc{CoT-ph}}\xspace}

\makeatletter
\def\bstctlcite{\@ifnextchar[{\@bstctlcite}{\@bstctlcite[@auxout]}}
\def\@bstctlcite[#1]#2{\@bsphack
  \@for\@citeb:=#2\do{%
    \edef\@citeb{\expandafter\@firstofone\@citeb}%
    \if@filesw\immediate\write\csname #1\endcsname{\string\citation{\@citeb}}\fi}%
  \@esphack}
\makeatother

\begin{document}
\bstctlcite{IEEEexample:BSTcontrol_custom}

\maketitle

\begin{abstract}

    We propose a Speech-to-Text Translation (S2TT) approach that integrates phoneme representations into a Chain-of-Thought (CoT) framework to improve translation in low-resource and zero-resource settings. By introducing phoneme recognition as an intermediate step, we enhance cross-lingual transfer, enabling translation even for languages with no labeled speech data. Our system builds on a multilingual LLM, which we extend to process speech and phonemes. Training follows a curriculum learning strategy that progressively introduces more complex tasks. Experiments on multilingual S2TT benchmarks show that phoneme-augmented CoT improves translation quality in low-resource conditions and enables zero-resource translation, while slightly impacting high-resource performance. Despite this trade-off, our findings demonstrate that phoneme-based CoT is a promising step toward making S2TT more accessible across diverse languages.

\end{abstract}

\begingroup
\renewcommand\thefootnote{$\dagger$}
\footnotetext{These authors contributed equally to this work.}
\endgroup

\section{Introduction}

Speech-to-Text Translation (S2TT) aims to convert speech in one language into text in another. Traditionally, these systems have relied on cascaded architectures, where an Automatic Speech Recognition (ASR) module first transcribes the speech, and a Text-to-Text Translation (T2TT) model translates the resulting text~\cite{waibel_spoken_2008}. While this modular approach eliminates the need for specialized S2TT datasets, it suffers from error propagation, as the T2TT stage cannot correct transcription errors. In contrast, direct S2TT models, or end-to-end~\cite{berard_listen_2016}, translate speech directly into the target language and have access to additional cues, such as prosody, which can be valuable for translation~\cite{tsiamas_speech_2024}. However, collecting parallel speech-text data across languages remains challenging. To address data scarcity, multitasking approaches are often adopted, where a single model learns ASR, T2TT, and S2TT jointly, leveraging large scale datasets~\cite{seamless_communication_team_joint_2025}.

Despite progress in reducing the need for dedicated S2TT datasets~\cite{duquenne_t-modules_2022}, obtaining sufficient labeled speech data remains a challenge for many languages. Large-scale collection efforts such as Common Voice~\cite{ardila_common_2020} have expanded multilingual speech resources, but some languages still have only a few hours of annotated recordings. In low-resource settings, this scarcity makes it difficult to train S2TT models, and in extreme cases, no labeled speech data is available at all. While some studies have explored strategies to mitigate this~\cite{mundnich_zero-resource_2025}, research remains limited.

In parallel, recent advances in Large Language Models (LLMs)~\cite{anil_palm_2023,openai_gpt-4_2024} have driven the development of unified architectures that integrate acoustic and textual processing. Speech Large Language Models (SLLMs)~\cite{tang_salmonn_2023,zhang_speechgpt_2023,rubenstein_audiopalm_2023,hassid_textually_2023,hu_wavllm_2024} extend pretrained LLMs to process spoken inputs by incorporating self-supervised speech encoders~\cite{hsu_hubert_2021}. These encoders transform speech into continuous representations, which are often discretized into tokens for seamless integration with language modeling. This approach allows SLLMs to perform speech-related tasks while leveraging the strong reasoning and contextual understanding of LLMs. As a result, SLLMs have been increasingly adopted in S2TT research, exploiting the underlying T2TT abilities of LLMs~\cite{wu_decoder-only_2023,huang_investigating_2024,chen_salm_2024}.

Inspired by the success of chain-of-thought (CoT) prompting in text-based reasoning, recent studies have explored its application to S2TT~\cite{huang_speech_2023,hu_chain--thought_2024,du_cot-st_2024}. This approach (S2TT-CoT) decomposes translation into intermediate steps, first generating a transcription and then producing the translation. While similar to cascaded systems, CoT preserves direct access to the audio during translation, reducing error accumulation and leveraging speech information. Notably,~\cite{huang_speech_2023} reports that CoT prompting outperforms cascaded inference on models trained on the same large-scale dataset. Additionally, CoT enables training with widely available ASR and T2TT data by explicitly representing sub-tasks of the translation process.

However, while many languages have parallel text data for T2TT, the lack of sufficient speech resources still hinders S2TT performance. To address this gap, we propose incorporating a Phoneme Recognition (PR) step into the CoT pipeline. Unlike ASR transcriptions, phonemes provide language-agnostic representations and can be generated from text, bridging the gap between speech and written forms. By training the model to recognize phonemes across multiple languages, we reduce reliance on transcript annotations and enable speech understanding even in languages never encountered in audio form.

In this work, we introduce \ourmodel, an LLM-based S2TT system that leverages CoT prompting to decompose the translation process. We extend this framework by incorporating an initial phoneme recognition step. Through this approach, we reduce reliance on speech data and enable S2TT even in languages without direct speech supervision. We evaluate our method in a multilingual setting, focusing on translation from multiple source languages into English. Our results demonstrate that CoT with phonemes improves translation quality in low-resource and zero-resource conditions.

\section{Methodology}

\begin{figure}[t]
  \centering
  \includegraphics[width=0.95\linewidth]{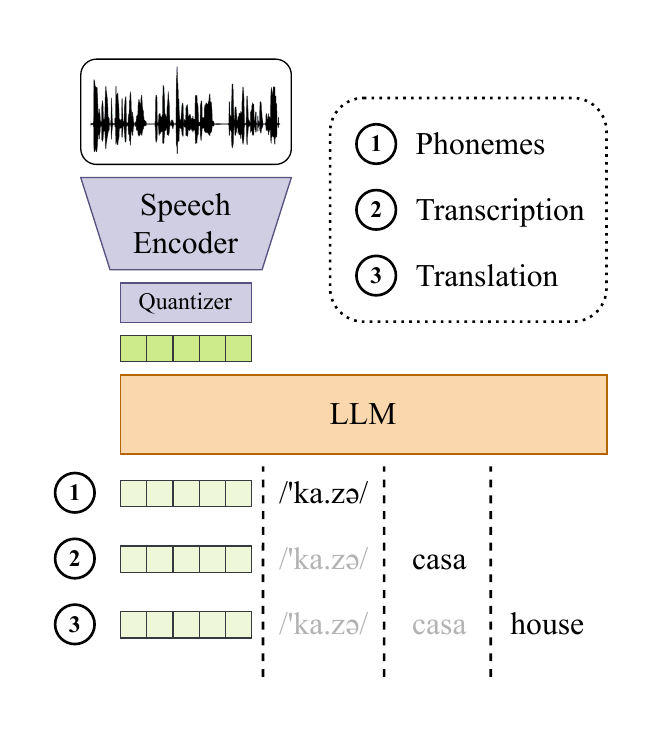}
  \vspace{-12pt}
  \caption{System overview and phoneme-augmented CoT.}
  \label{fig:arch}
  \vspace{-12pt}
\end{figure}

Our system, \ourmodel, is an LLM-based S2TT model built upon an open-source multilingual text-only LLM~\cite{gonzalez-agirre_salamandra_2025}. We extend it to S2TT by incorporating speech tokens as inputs and training it to perform CoT, following~\cite{huang_speech_2023}. An overview of the system is shown in Fig. \ref{fig:arch}, with full architectural details provided in \S\ref{sec:arch}. Additionally, we enhance the model by integrating phoneme representations (\S\ref{sec:phonemes}) and adopting a curriculum learning approach that progressively increases task complexity during training (\S\ref{sec:cl}).

\subsection{Architecture}
\label{sec:arch}

Following recent work~\cite{zhang_speechgpt_2023,rubenstein_audiopalm_2023}, we extend the base LLM to handle speech inputs. We encode speech using a self-supervised pretrained model \( f_{\text{enc}} \)~\cite{hsu_hubert_2021}, which extracts continuous representations from an input utterance \( x \). These continuous features are then quantized into a sequence of discrete tokens:
\(
\bm{s} = (s_1, \ldots, s_T) = q\bigl(f_{\text{enc}}(x)\bigr),\quad s_t \in V_s
\),
where \( q(\cdot) \) is the quantization function and \( V_s \) is the speech tokens vocabulary.

In addition to speech tokens, we incorporate phoneme tokens to favor cross-lingual speech understanding (\S\ref{sec:phonemes}). To accommodate both input types, we expand the model's vocabulary and embedding matrix. Given an original vocabulary \( V_o \) of size \( |V_o| \), we introduce the speech tokens vocabulary \( V_s \) and the phonemes vocabulary \( V_p \). The expanded vocabulary is:
\(
V = V_o \cup V_s \cup V_p
\).
Its embedding matrix is updated to:
\(
E = \bigl[E_o;\,E_s;\,E_p\bigr] \in \mathbb{R}^{(|V|)\times d}
\),
where \( E_o \) retains the original embeddings, and \( E_s \) and \( E_p \) are randomly initialized.

\subsection{CoT with Phoneme Recognition}
\label{sec:phonemes}

To enhance cross-lingual transferability and reduce the need for ASR data for each supported language, we extend the standard S2TT-CoT approach by incorporating a Phoneme Recognition (PR) step. Unlike conventional ASR transcriptions, phonemes provide a language-agnostic representation that does not depend on a language's writing system. Instead of directly transcribing speech into text, our model first predicts phonemes, maps them to a transcription, and then translates the text into English. This three-step CoT process is illustrated in Fig.~\ref{fig:arch}, and an example prompt format is shown in Fig.~\ref{fig:prompt}.

We generate phoneme sequences synthetically from the transcriptions of ASR and S2TT datasets, as well as from the source side of T2TT corpora. The speech-to-phoneme mapping is expected to generalize across languages, allowing the model to produce phonemes even for languages with little or no available speech data. In parallel, we instruct the model to translate phonemes from all supported languages into English, bridging the gap between speech inputs and transcriptions. As in the original S2TT-CoT approach, our method remains compatible with large-scale ASR and T2TT training data.

\subsection{Curriculum Learning}
\label{sec:cl}

We adopt a curriculum learning approach, similar to strategies applied in SLLMs~\cite{zhang_speechgpt_2023,tang_salmonn_2023,hu_wavllm_2024} and S2TT~\cite{huang_speech_2023,du_cot-st_2024}. Our curriculum consists of three stages, each designed to gradually introduce new capabilities. The model first aligns its new representations, then learns individual tasks, and finally integrates them into a multi-step CoT framework.

\noindent\textbf{Stage 1}\;~The first stage focuses on integrating newly added speech tokens and phonemes into the latent space of the base LLM. After expanding the model's vocabulary (\S\ref{sec:arch}), we train it on next-token prediction, first with speech tokens and then with phonemes. This follows an approach similar to SpeechGPT~\cite{tang_salmonn_2023}, except that we train only the embedding and output projection layers, keeping the backbone frozen. This allows the new embeddings to adapt to the base LLM representation space without affecting its existing capabilities. We have found that this initialization phase accelerates convergence in later stages and benefits downstream task performance.

\noindent\textbf{Stage 2}\;~Next, we train the entire model in a multitask setup consisting of Phoneme Recognition (PR), Phoneme-to-Grapheme (P2G), Grapheme-to-Phoneme (G2P), Automatic Speech Recognition (ASR), and Text-to-Text Translation (T2TT). This stage ensures that the model learns the fundamental tasks required to perform the full CoT process. We do not train on Speech-to-Text Translation (S2TT) at this stage, following the curriculum learning idea of starting with simpler tasks before progressing to more complex ones~\cite{hu_wavllm_2024}.

\noindent\textbf{Stage 3}\;~In the final stage, we fine-tune the model to perform S2TT with a CoT approach, with intermediate steps on phonemes and transcriptions (S2TT-CoT-Ph). We also keep leveraging ASR and T2TT datasets in this stage, but in this case expanded with phonemes. We do this by training subchains of the full one, whether from speech to the transcription, or from phonemes in one language to graphemes in English (both with the corresponding intermediate step). We refer to these tasks as ASR-CoT and Phoneme-to-Text Translation (P2TT-CoT).

\section{Experimental Setup}

\subsection{Data}
\label{sec:data}

Our training framework includes multiple tasks across different languages and modalities (\S\ref{sec:cl}). This requires careful selection of training data from various sources. Below, we describe our data composition, language selection, and statistical analysis.

\noindent\textbf{Datasets}\;~We use a combination of datasets for each task in our setup. For ASR, we train on Common Voice 17.0 (CV17)~\cite{ardila_common_2020} and the supervised portion of VoxPopuli~\cite{wang_voxpopuli_2021}. We also leverage these datasets to generate training data for the PR task using \texttt{eSpeak}.\footnote{\url{https://github.com/espeak-ng/espeak-ng}} For T2TT, we use data from NLLB~\cite{costa-jussa_scaling_2024}. Given the massive size of this dataset, we subsample it by selecting the highest-quality sentence pairs based on alignment scores from LASER3~\cite{heffernan_bitext_2022}. We then split the data into non-overlapping chunks across tasks and stages: the top $375k$ samples are used for T2TT/P2TT training in the final stage, and the next $375k$ for the same tasks in stage 2. For S2TT, we train on CoVoST 2~\cite{wang_covost_2021}. We conduct our evaluation on FLEURS~\cite{conneau_fleurs_2023}. We omit CoVoST 2 for evaluation due to significant overlap between its test split and Common Voice training data.

\noindent\textbf{Languages}\;~For choosing the source speech languages, we prioritize phonetically similar ones and focus on European languages. Our study includes the Germanic (Dutch, German, Swedish), Romance (Catalan, Italian, Spanish), and Slavic (Polish, Russian, Slovenian) language groups. To examine cross-lingual transfer to languages under zero-resource settings, we exclude one language from each group from all speech training data (ASR/PR and S2TT), selecting Dutch, Italian, and Polish. In total, we consider 9 source languages, translating exclusively into English.

\noindent\textbf{Statistics}\;~We use approximately $8.5k$ hours of speech-transcription pairs and a total of $455$ hours for S2TT (Table~\ref{tab:hours_per_dataset}). English is used exclusively for ASR. Catalan, German, and Spanish have the most speech data and are thus considered under high-resource conditions. Russian has only $18$ hours of S2TT data but benefits from a larger CV corpus, so we classify it as a mid-resource scenario. Swedish and Slovenian have just $2$ hours of supervised S2TT data and fewer than $50$ total speech hours, placing them under low-resource conditions. For T2TT, we use the same amount of data for each language, so there is no language imbalance.

\begin{table}[t]
    \centering
    \small
    \renewcommand{\arraystretch}{1.3}
    \setlength{\tabcolsep}{4pt}
    \caption{\textit{Number of speech hours used for training.}}
    \begin{tabular}{lccccccc}
        \toprule
        \textbf{Dataset} & \textbf{En} & \textbf{Ca} & \textbf{De} & \textbf{Es} & \textbf{Ru} & \textbf{Sv} & \textbf{Sl} \\
        \midrule
        CV17~\cite{ardila_common_2020} & 2615 & 2698 & 1334 & 562 & 235 & 46 & 23 \\
        VoxPopuli~\cite{wang_voxpopuli_2021} & 543 & - & 282 & 166 & - & - & 10 \\
        CoVoST 2~\cite{wang_covost_2021} & - & 136 & 184 & 113 & 18 & 2 & 2 \\
        \midrule
        Total & 3158 & 2834 & 1800 & 841 & 253 & 48 & 35 \\
        \bottomrule
    \end{tabular}
    \label{tab:hours_per_dataset}
    \vspace{-9pt}
\end{table}

\begin{table*}[ht]
\centering
\scriptsize
\setlength{\tabcolsep}{3pt} 
\renewcommand{\arraystretch}{1.1} 
\caption{BLEU and COMET scores on the FLEURS test set (xx$\rightarrow$en). Both metrics have been computed with paired bootstrap resampling to obtain statistical significance with respect to \ourmodelcot. The \underline{underlined} values correspond to scores with a $\text{p-value}<0.05$. Superscripts indicate CoT generation with (\textdaggerdbl) and without (\textdagger) phonemes at inference.}
\begin{tabular}{l *{12}{r} @{\hskip 10pt} *{12}{r}}
\toprule
  & \multicolumn{12}{c}{BLEU ($\uparrow$)} & \multicolumn{12}{c}{COMET ($\uparrow$)} \\
\cmidrule(lr){2-13} \cmidrule(lr){14-25}
  & \multicolumn{4}{c}{High-Resource} 
  & \multicolumn{4}{c}{Mid-/Low-Resource} 
  & \multicolumn{4}{c}{Zero-Resource} 
  & \multicolumn{4}{c}{High-Resource} 
  & \multicolumn{4}{c}{Mid-/Low-Resource} 
  & \multicolumn{4}{c}{Zero-Resource} \\
\cmidrule(lr){2-5}  \cmidrule(lr){6-9}  \cmidrule(lr){10-13}
\cmidrule(lr){14-17} \cmidrule(lr){18-21} \cmidrule(lr){22-25}
Model 
  & \multicolumn{1}{c}{Ca} & \multicolumn{1}{c}{De} & \multicolumn{1}{c}{Es} & \multicolumn{1}{c}{Avg.} 
  & \multicolumn{1}{c}{Ru} & \multicolumn{1}{c}{Sv} & \multicolumn{1}{c}{Sl} & \multicolumn{1}{c}{Avg.} 
  & \multicolumn{1}{c}{It} & \multicolumn{1}{c}{Nl} & \multicolumn{1}{c}{Pl} & \multicolumn{1}{c}{Avg.} 
  & \multicolumn{1}{c}{Ca} & \multicolumn{1}{c}{De} & \multicolumn{1}{c}{Es} & \multicolumn{1}{c}{Avg.} 
  & \multicolumn{1}{c}{Ru} & \multicolumn{1}{c}{Sv} & \multicolumn{1}{c}{Sl} & \multicolumn{1}{c}{Avg.} 
  & \multicolumn{1}{c}{It} & \multicolumn{1}{c}{Nl} & \multicolumn{1}{c}{Pl} & \multicolumn{1}{c}{Avg.} \\
\midrule
\textsc{direct} 
  & \underline{28.3}        & \underline{26.6}        & \underline{18.0}        & 24.3        
  & \underline{9.5}         & \underline{6.6}         & \underline{6.1}       & 7.4         
  & \underline{8.4}         & \underline{4.8}         & \underline{2.7}         & 5.3        
  & \underline{79.8} & \underline{79.6} & \underline{78.8} & 79.4 
  & 61.4 & \underline{53.6} & 56.3 & 57.1 
  & \underline{61.2} & \underline{\textbf{53.2}} & \underline{49.0} & 54.5 \\
\textsc{CoT}\textsuperscript{\textdagger}    
  & 34.0 & \textbf{31.5} & \textbf{21.4} & \textbf{29.0} 
  & 13.1 & 13.5 & 9.1  & 11.9 
  & 9.3  & 3.3  & 1.8    & 4.8 
  & \textbf{82.9} & \textbf{83.0} & \textbf{81.4} & \textbf{82.4} 
  & 61.0 & 57.0 & 56.2 & 58.1
  & 58.9 & 43.9 & 39.1 & 47.3 \\
\midrule
\textsc{CoT-ph}\textsuperscript{\textdaggerdbl} 
  & \underline{32.5}        & \underline{29.2}        & \underline{19.9}        & 27.2        
  & 13.4        & 14.0        & \underline{10.4}       & 12.6        
  & \underline{11.0}        & 2.9         & 1.6          & 5.2        
  & \underline{80.2} & \underline{80.0} & \underline{77.9} & 79.4 
  & \underline{59.5} & \underline{55.2} & 56.9 & 57.2 
  & 59.6 & 43.6 & \underline{43.8} & 49.0 \\
\quad + PDA\textsuperscript{\textdaggerdbl}     
  & \underline{33.0}        & \underline{29.5}        & \underline{20.7}        & 27.7        
  & \underline{14.2} & \underline{15.7} & \underline{12.2}      & 14.0 
  & \underline{13.6}        & \underline{5.5} & 2.2          & 7.1        
  & \underline{80.5} & \underline{80.9} & \underline{79.4} & 80.3 
  & 61.9 & 57.2 & \underline{60.3} & 59.8
  & \underline{64.6} & \underline{48.1} & \underline{45.2} & 52.6 \\
\quad + DPS\textsuperscript{\textdaggerdbl}     
  & \underline{33.2}           & \underline{30.4}           & \underline{20.8}           & 28.1           
  & 13.8           & \underline{14.9}           & \underline{12.9}         & 13.9           
  & \underline{13.7}           & \underline{5.7}           & \underline{3.0}          & 7.5           
  & \underline{80.7} & \underline{80.7} & \underline{79.5} & 80.3 
  & 61.2 & 57.2 & \underline{59.6} & 59.3 
  & \underline{65.1} & \underline{49.1} & \underline{46.9} & 53.7 \\
\quad + DPS\textsuperscript{\textdagger}     
  & \textbf{34.5} & \underline{30.0} & \underline{20.9} & 28.5 
  & \underline{\textbf{16.2}} & \underline{\textbf{18.5}} & \underline{\textbf{14.6}} & \textbf{16.4} 
  & \underline{\textbf{15.5}} & \underline{\textbf{7.9}} & \underline{\textbf{4.7}} & \textbf{9.4} 
  & 82.5& \underline{81.6}& \underline{80.8}&  81.6          
  & \underline{\textbf{65.9}}& \underline{\textbf{62.2}}& \underline{\textbf{63.8}}& \textbf{64.0}  
  & \underline{\textbf{68.3}}& \underline{53.1}& \underline{\textbf{49.3}}& \textbf{56.9} \\
\bottomrule
\end{tabular}
\label{tab:results}
\vspace{-9pt}
\end{table*}

\subsection{Implementation Details}

\noindent\textbf{Speech Encoder}\;~We encode speech utterances with mHuBERT from TWIST~\cite{hassid_textually_2023}, using \texttt{textless-lib}~\cite{kharitonov_textless-lib_2022}. We chose this module for two reasons: (1) it is capable of encoding multilingual speech, and (2) it reduces the sample rate to 25 Hz, halving the temporal resolution of most common HuBERTs~\cite{hsu_hubert_2021}. We extract discrete speech tokens by clustering the \nth{11}-layer representations into 500 groups. 

\noindent\textbf{Base LLM}\;~The main component of our system is Salamandra 2B~\cite{gonzalez-agirre_salamandra_2025}, a highly multilingual LLM trained on 35 European languages and code. We chose this model because we particularly want to cover language families from Europe, as described in \S\ref{sec:data}. Specifically, we used its instructed version,\footnote{\url{https://hf.co/BSC-LT/salamandra-2b-instruct}} so all the data we use is formatted with the corresponding chat template based on OpenAI's ChatML.

\noindent\textbf{Hyperparameters}\;~We train all stages (\S\ref{sec:cl}) using the AdamW optimizer~\cite{loshchilov_decoupled_2019}, a cosine learning rate scheduler, and gradient clipping with a maximum norm of $1.0$. Each stage has slight variations in the setup. In stage 1, we use a learning rate of $7 \cdot 10^{-5}$, a per-device batch size of $64$, and four GPUs, resulting in an effective batch size of $256$. We train for one epoch and warm up the learning rate during the first $3\%$ of updates. We set the maximum sequence length to $1024$ and pack tokens from different sequences into full-length sequences to improve training efficiency, following~\cite{zhang_speechgpt_2023}. In later stages, we increase the effective batch size to $512$ by training on $16$ GPUs while reducing the per-device batch size to $32$. In stage 2, we train for two epochs, lower the learning rate to $4 \cdot 10^{-5}$, and warm it up over the first $10\%$ of updates. We maintain the same sequence length but without packing sequences. In stage 3, we extend the maximum sequence length to $2048$ to accommodate full CoT samples, further reduce the learning rate to $1 \cdot 10^{-5}$, and train for one epoch. We use the \texttt{transformers} library~\cite{wolf_transformers_2020} and handle distributed training with DeepSpeed. All experiments run on NVIDIA H100 GPUs. To optimize memory usage, we apply mixed precision (bfloat16) and gradient checkpointing. We leverage Liger Kernel~\cite{hsu_liger_2025} to further improve memory efficiency and increase training throughput.

\noindent\textbf{Inference Strategy}\;~We implement a controlled generation instead of allowing free-form CoT generation to prevent potential divergence in the steps. This process follows a three-stage pipeline, corresponding to the generation of phonemes, transcription, and translation. Between generation steps, we append the newly generated tokens and task-specific prompts to guide the subsequent generation phase. With this approach, we aim to ensure consistent and reliable output generation. In all generation steps we use beam-search multinomial sampling with early stopping. We set a temperature of 0.2, a top-p of 0.95, and a top-k of 50. When generating phonemes we reduce the top-k to 10, to compensate for the smaller vocabulary.

\begin{figure}[t]
    \centering
    \begin{tcolorbox}[
        colback=gray!8, 
        colframe=gray!127, 
        boxrule=0.8pt, 
        title={Prompt Format}, 
        coltitle=white, 
        colbacktitle=gray!127, 
    ]

    {\normalsize \textbf{Input}}

    \vspace{3pt}

    Convert the following speech audio into a phonemic sequence in \{language\}, then transcribe it to graphemes and then translate it to English. The audio is:\\
    {\small \texttt{<sosp><409><7><...><87><409><eosp>}}

    \vspace{9pt}

    {\normalsize \textbf{Output}}

    \vspace{3pt}

    The phonemic sequence in \{language\} is:\\
    \textbf{\{phonemes\}}\\

    The transcription of the phonemic sequence in \{language\} is:\\
    \textbf{\{transcription\}}\\

    The translation from \{language\} to English is:
    \textbf{\{translation\}}
    \end{tcolorbox}
    \caption{Phoneme-Augmented CoT format.}
    \label{fig:prompt}
    \vspace{-12pt}
\end{figure}

\subsection{Experiments}

We evaluate our approach by comparing it against two baselines. We also explore two complementary strategies designed to mitigate the limitations of our method. The baselines are:

\noindent\textbf{\ourmodeldirect}\;~This model translates speech directly without intermediate steps. We train it on the ASR and T2TT tasks during stage 2, and add S2TT (without CoT) in the last stage (see \S\ref{sec:cl}).

\noindent\textbf{\ourmodelcot}\;~This model translates speech following the CoT prompting strategy, without phoneme recognition. We train it on the same ASR and T2TT tasks in stage 2, and add S2TT-CoT in stage 3.

We refer to our main model as \textbf{\ourmodelphcot} (see Figure~\ref{fig:prompt} and \S\ref{sec:phonemes}), and we introduce two refined variations:

\noindent\textbf{Phoneme Data Augmentation (PDA)}\;~This variant introduces a regularization mechanism in the phoneme recognition step of the ASR-CoT and S2TT-CoT-Ph tasks in stage 3. Its goal is to encourage the model to rely on both audio and phonemes, mitigating error propagation issues observed in preliminary analysis. The augmentation process randomly deletes, masks, or substitutes phoneme spans, inserts random phonemes, and shifts spaces. To preserve the phoneme capabilities acquired in stage 2, we keep 25\% unchanged, and omit training the phoneme step on the modified samples.

\noindent\textbf{Dual Prompting Strategy (DPS)}\;~This variant enables the model to operate without the phoneme recognition step at inference time while still leveraging it during training. To support this dual behavior, we modify the training mix: 20\% of samples exclude the phonemes step, while the remaining is 75\% of PDA data and 5\% of original samples.

\section{Results}

In Table \ref{tab:results}, we report the BLEU~\cite{papineni_bleu_2002} and COMET~\cite{rei_comet_2020} scores on FLEURS~\cite{conneau_fleurs_2023}. Analyzing the results across language families and resource scenarios, we confirm the following findings:

\noindent\textbf{CoT improves Direct}\;~\ourmodelcot consistently outperforms \ourmodeldirect in all non-zero-resource settings, with average gains over 4.5 BLEU. This confirms the effectiveness of S2TT-CoT, which underpins our method.

\noindent\textbf{Phonemes help low-resource}\;~\ourmodelphcot improves performance in mid-/low-resource, and zero-resource settings, but degrades it in high-resource scenarios. In mid-/low-resource settings, it achieves an average gain of 0.7 BLEU over \ourmodelcot, driven by improvements in Swedish and Slovenian. Conversely, in high-resource settings, incorporating phonemes reduces performance by an average of 1.8 BLEU points. These results suggest that phonemes help bridge the modality gap between speech and written forms. Although the phoneme recognition step introduces a potential source of errors, it provides a net benefit in low-resource conditions. Beyond a certain data threshold, however, its advantages diminish and it may instead harm translation quality.

\noindent\textbf{PDA enhances robustness}\;~This technique amplifies the benefits of our method and narrows the performance gap in high-resource settings. It yields average gains of 2.1 BLEU in mid-/low-resource scenarios, and 2.3 in zero-resource conditions over \ourmodelcot. Degradation in high-resource settings is reduced, with performance now declining by 1.3 BLEU points on average. These results show that PDA improves robustness by reducing sensitivity to phoneme recognition errors.

\noindent\textbf{DPS yields strongest results}\;~DPS enables flexible decoding, allowing generation with or without phonemes. Notably, after training with DPS, generating \textit{without} the phoneme recognition step (DPS\textsuperscript{\textdagger}) yields the strongest results, with average gains above 4.5 BLEU points over \ourmodelcot across mid-/low-resource, and zero-resource settings. In high-resource scenarios, the performance gap is nearly closed, with only a 0.5 BLEU degradation on average. In contrast, decoding \textit{with} phonemes (DPS\textsuperscript{\textdaggerdbl}) performs close to PDA. These findings suggest that phonemes are particularly beneficial during training, as they facilitate cross-modal alignment, while skipping them at inference helps avoid error propagation issues.

\noindent\textbf{Cross-lingual transfer depends on language proximity}\;~Zero-resource performance is influenced by the availability of data in related languages. Italian, for instance, benefits from nearby high-resource Catalan and Spanish, achieving the best zero-resource results overall. In contrast, Polish reaches only 4.7 BLEU, likely due to limited availability of training data in other Slavic languages. These patterns highlight the interaction between linguistic similarity and data availability. Our methods help mitigate data scarcity but cannot fully bridge structural gaps between distant languages.

\section{Conclusions}

This work introduces a phoneme-augmented CoT framework for S2TT, demonstrating its effectiveness in low-resource and zero-resource settings. By incorporating phoneme recognition as an intermediate step, we enhance cross-lingual transfer and improve translation performance when labeled speech data are scarce. Our results show that while including phonemes benefits low-resource scenarios, it slightly degrades in high-resource settings. Additionally, we also propose a dual prompting strategy that allows flexible inference with or without phoneme recognition. Surprisingly, this also brought benefits even if used without phonemes at inference time. These findings highlight the potential of phoneme representation for improving multilingual S2TT, particularly for underrepresented languages. Future research should focus on further reducing error propagation in multi-step inference, and handling accent variability.

\section{Acknowledgements}

This work is funded by the Ministerio para la Transformación Digital y de la Función Pública and Plan de Recuperación, Transformación y Resiliencia -- Funded by EU -- NextGenerationEU within the framework of the project Modelos del Lenguaje. 

\newpage

\bibliographystyle{IEEEtran_noeditor}
\bibliography{\jobname, references}

\begin{thebibliography}{10}
\providecommand{\url}[1]{#1}
\csname url@samestyle\endcsname
\providecommand{\newblock}{\relax}
\providecommand{\bibinfo}[2]{#2}
\providecommand{\BIBentrySTDinterwordspacing}{\spaceskip=0pt\relax}
\providecommand{\BIBentryALTinterwordstretchfactor}{4}
\providecommand{\BIBentryALTinterwordspacing}{\spaceskip=\fontdimen2\font plus
\BIBentryALTinterwordstretchfactor\fontdimen3\font minus \fontdimen4\font\relax}
\providecommand{\BIBforeignlanguage}[2]{{%
\expandafter\ifx\csname l@#1\endcsname\relax
\typeout{** WARNING: IEEEtran.bst: No hyphenation pattern has been}%
\typeout{** loaded for the language `#1'. Using the pattern for}%
\typeout{** the default language instead.}%
\else
\language=\csname l@#1\endcsname
\fi
#2}}
\providecommand{\BIBdecl}{\relax}
\BIBdecl

\bibitem{waibel_spoken_2008}
\BIBentryALTinterwordspacing
A.~Waibel and C.~Fugen, ``Spoken language translation,'' \emph{IEEE Signal Processing Magazine}, vol.~25, no.~3, pp. 70--79, May 2008, conference Name: IEEE Signal Processing Magazine.
\BIBentrySTDinterwordspacing

\bibitem{berard_listen_2016}
\BIBentryALTinterwordspacing
A.~Bérard, O.~Pietquin, L.~Besacier \emph{et~al.}, ``Listen and {Translate}: {A} {Proof} of {Concept} for {End}-to-{End} {Speech}-to-{Text} {Translation},'' in \emph{{NIPS} {Workshop} on end-to-end learning for speech and audio processing}, Barcelona, Spain, Dec. 2016.
\BIBentrySTDinterwordspacing

\bibitem{tsiamas_speech_2024}
\BIBentryALTinterwordspacing
I.~Tsiamas, M.~Sperber, A.~Finch \emph{et~al.}, ``Speech {Is} {More} than {Words}: {Do} {Speech}-to-{Text} {Translation} {Systems} {Leverage} {Prosody}?'' in \emph{Proceedings of the {Ninth} {Conference} on {Machine} {Translation}}.\hskip 1em plus 0.5em minus 0.4em\relax Miami, Florida, USA: Association for Computational Linguistics, Nov. 2024, pp. 1235--1257.
\BIBentrySTDinterwordspacing

\bibitem{seamless_communication_team_joint_2025}
\BIBentryALTinterwordspacing
{SEAMLESS Communication Team}, ``\BIBforeignlanguage{en}{Joint speech and text machine translation for up to 100 languages},'' \emph{\BIBforeignlanguage{en}{Nature}}, vol. 637, no. 8046, pp. 587--593, Jan. 2025, publisher: Nature Publishing Group.
\BIBentrySTDinterwordspacing

\bibitem{duquenne_t-modules_2022}
\BIBentryALTinterwordspacing
P.-A. Duquenne, H.~Gong, B.~Sagot \emph{et~al.}, ``T-{Modules}: {Translation} {Modules} for {Zero}-{Shot} {Cross}-{Modal} {Machine} {Translation},'' in \emph{Proceedings of the 2022 {Conference} on {Empirical} {Methods} in {Natural} {Language} {Processing}}.\hskip 1em plus 0.5em minus 0.4em\relax Abu Dhabi, United Arab Emirates: Association for Computational Linguistics, Dec. 2022, pp. 5794--5806.
\BIBentrySTDinterwordspacing

\bibitem{ardila_common_2020}
\BIBentryALTinterwordspacing
R.~Ardila, M.~Branson, K.~Davis \emph{et~al.}, ``\BIBforeignlanguage{eng}{Common {Voice}: {A} {Massively}-{Multilingual} {Speech} {Corpus}},'' in \emph{\BIBforeignlanguage{eng}{Proceedings of the {Twelfth} {Language} {Resources} and {Evaluation} {Conference}}}.\hskip 1em plus 0.5em minus 0.4em\relax Marseille, France: European Language Resources Association, May 2020, pp. 4218--4222.
\BIBentrySTDinterwordspacing

\bibitem{mundnich_zero-resource_2025}
\BIBentryALTinterwordspacing
K.~Mundnich, X.~Niu, P.~Mathur \emph{et~al.}, ``Zero-resource {Speech} {Translation} and {Recognition} with {LLMs},'' in \emph{{ICASSP} 2025 - 2025 {IEEE} {International} {Conference} on {Acoustics}, {Speech} and {Signal} {Processing} ({ICASSP})}, Apr. 2025, pp. 1--5.
\BIBentrySTDinterwordspacing

\bibitem{anil_palm_2023}
\BIBentryALTinterwordspacing
R.~Anil, A.~M. Dai, O.~Firat \emph{et~al.}, ``{PaLM} 2 {Technical} {Report},'' 2023, version Number: 3.
\BIBentrySTDinterwordspacing

\bibitem{openai_gpt-4_2024}
\BIBentryALTinterwordspacing
OpenAI, J.~Achiam, S.~Adler \emph{et~al.}, ``{GPT}-4 {Technical} {Report},'' Mar. 2024, arXiv:2303.08774 [cs].
\BIBentrySTDinterwordspacing

\bibitem{tang_salmonn_2023}
\BIBentryALTinterwordspacing
C.~Tang, W.~Yu, G.~Sun \emph{et~al.}, ``\BIBforeignlanguage{en}{{SALMONN}: {Towards} {Generic} {Hearing} {Abilities} for {Large} {Language} {Models}},'' in \emph{\BIBforeignlanguage{en}{The {Twelfth} {ICLR}}}, Oct. 2023.
\BIBentrySTDinterwordspacing

\bibitem{zhang_speechgpt_2023}
\BIBentryALTinterwordspacing
D.~Zhang, S.~Li, X.~Zhang \emph{et~al.}, ``{SpeechGPT}: {Empowering} {Large} {Language} {Models} with {Intrinsic} {Cross}-{Modal} {Conversational} {Abilities},'' in \emph{Findings of the {Association} for {Computational} {Linguistics}: {EMNLP} 2023}.\hskip 1em plus 0.5em minus 0.4em\relax Singapore: Association for Computational Linguistics, Dec. 2023, pp. 15\,757--15\,773.
\BIBentrySTDinterwordspacing

\bibitem{rubenstein_audiopalm_2023}
\BIBentryALTinterwordspacing
P.~K. Rubenstein, C.~Asawaroengchai, D.~D. Nguyen \emph{et~al.}, ``\BIBforeignlanguage{en}{{AudioPaLM}: {A} {Large} {Language} {Model} {That} {Can} {Speak} and {Listen}},'' Jun. 2023, arXiv:2306.12925 [cs].
\BIBentrySTDinterwordspacing

\bibitem{hassid_textually_2023}
\BIBentryALTinterwordspacing
M.~Hassid, T.~Remez, T.~A. Nguyen \emph{et~al.}, ``\BIBforeignlanguage{en}{Textually {Pretrained} {Speech} {Language} {Models}},'' in \emph{\BIBforeignlanguage{en}{Thirty-seventh {Conference} on {Neural} {Information} {Processing} {Systems}}}, Nov. 2023.
\BIBentrySTDinterwordspacing

\bibitem{hu_wavllm_2024}
\BIBentryALTinterwordspacing
S.~Hu, L.~Zhou, S.~Liu \emph{et~al.}, ``{WavLLM}: {Towards} {Robust} and {Adaptive} {Speech} {Large} {Language} {Model},'' in \emph{Findings of the {Association} for {Computational} {Linguistics}: {EMNLP} 2024}.\hskip 1em plus 0.5em minus 0.4em\relax Miami, Florida, USA: Association for Computational Linguistics, Nov. 2024, pp. 4552--4572.
\BIBentrySTDinterwordspacing

\bibitem{hsu_hubert_2021}
\BIBentryALTinterwordspacing
W.-N. Hsu, B.~Bolte, Y.-H.~H. Tsai \emph{et~al.}, ``{HuBERT}: {Self}-{Supervised} {Speech} {Representation} {Learning} by {Masked} {Prediction} of {Hidden} {Units},'' \emph{IEEE/ACM Trans. Audio, Speech and Lang. Proc.}, vol.~29, pp. 3451--3460, Oct. 2021.
\BIBentrySTDinterwordspacing

\bibitem{wu_decoder-only_2023}
\BIBentryALTinterwordspacing
J.~Wu, Y.~Gaur, Z.~Chen \emph{et~al.}, ``\BIBforeignlanguage{en}{On decoder-only architecture for speech-to-text and large language model integration},'' Oct. 2023, arXiv:2307.03917 [eess].
\BIBentrySTDinterwordspacing

\bibitem{huang_investigating_2024}
\BIBentryALTinterwordspacing
C.-W. Huang, H.~Lu, H.~Gong \emph{et~al.}, ``Investigating {Decoder}-only {Large} {Language} {Models} for {Speech}-to-text {Translation},'' in \emph{Proc. {Interspeech} 2024}, 2024, pp. 832--836.
\BIBentrySTDinterwordspacing

\bibitem{chen_salm_2024}
\BIBentryALTinterwordspacing
Z.~Chen, H.~Huang, A.~Andrusenko \emph{et~al.}, ``{SALM}: {Speech}-{Augmented} {Language} {Model} with in-{Context} {Learning} for {Speech} {Recognition} and {Translation},'' in \emph{{ICASSP} 2024 - 2024 {IEEE} {International} {Conference} on {Acoustics}, {Speech} and {Signal} {Processing} ({ICASSP})}, Apr. 2024, pp. 13\,521--13\,525.
\BIBentrySTDinterwordspacing

\bibitem{huang_speech_2023}
\BIBentryALTinterwordspacing
Z.~Huang, R.~Ye, T.~Ko \emph{et~al.}, ``Speech {Translation} with {Large} {Language} {Models}: {An} {Industrial} {Practice},'' Dec. 2023, arXiv:2312.13585 [cs].
\BIBentrySTDinterwordspacing

\bibitem{hu_chain--thought_2024}
\BIBentryALTinterwordspacing
K.~Hu, Z.~Chen, C.-H.~H. Yang \emph{et~al.}, ``Chain-of-{Thought} {Prompting} for {Speech} {Translation},'' Sep. 2024, arXiv:2409.11538 [cs].
\BIBentrySTDinterwordspacing

\bibitem{du_cot-st_2024}
\BIBentryALTinterwordspacing
Y.~Du, Z.~Ma, Y.~Yang \emph{et~al.}, ``{CoT}-{ST}: {Enhancing} {LLM}-based {Speech} {Translation} with {Multimodal} {Chain}-of-{Thought},'' Sep. 2024, arXiv:2409.19510 [cs].
\BIBentrySTDinterwordspacing

\bibitem{gonzalez-agirre_salamandra_2025}
\BIBentryALTinterwordspacing
A.~Gonzalez-Agirre, M.~Pàmies, J.~Llop \emph{et~al.}, ``Salamandra {Technical} {Report},'' Feb. 2025, arXiv:2502.08489 [cs].
\BIBentrySTDinterwordspacing

\bibitem{wang_voxpopuli_2021}
\BIBentryALTinterwordspacing
C.~Wang, M.~Riviere, A.~Lee \emph{et~al.}, ``{VoxPopuli}: {A} {Large}-{Scale} {Multilingual} {Speech} {Corpus} for {Representation} {Learning}, {Semi}-{Supervised} {Learning} and {Interpretation},'' in \emph{Proceedings of the 59th {Annual} {Meeting} of the {ACL} and the 11th {IJCNLP} ({Volume} 1: {Long} {Papers})}.\hskip 1em plus 0.5em minus 0.4em\relax Online: Association for Computational Linguistics, Aug. 2021, pp. 993--1003.
\BIBentrySTDinterwordspacing

\bibitem{costa-jussa_scaling_2024}
\BIBentryALTinterwordspacing
M.~R. Costa-jussà, J.~Cross, O.~Çelebi \emph{et~al.}, ``\BIBforeignlanguage{en}{Scaling neural machine translation to 200 languages},'' \emph{\BIBforeignlanguage{en}{Nature}}, vol. 630, no. 8018, pp. 841--846, Jun. 2024, publisher: Nature Publishing Group.
\BIBentrySTDinterwordspacing

\bibitem{heffernan_bitext_2022}
\BIBentryALTinterwordspacing
K.~Heffernan, O.~Çelebi, and H.~Schwenk, ``Bitext {Mining} {Using} {Distilled} {Sentence} {Representations} for {Low}-{Resource} {Languages},'' in \emph{Findings of the {ACL}: {EMNLP} 2022}.\hskip 1em plus 0.5em minus 0.4em\relax Abu Dhabi, United Arab Emirates: Association for Computational Linguistics, Dec. 2022, pp. 2101--2112.
\BIBentrySTDinterwordspacing

\bibitem{wang_covost_2021}
\BIBentryALTinterwordspacing
C.~Wang, A.~Wu, J.~Gu \emph{et~al.}, ``{CoVoST} 2 and {Massively} {Multilingual} {Speech} {Translation},'' in \emph{Proc. {Interspeech} 2021}, 2021, pp. 2247--2251.
\BIBentrySTDinterwordspacing

\bibitem{conneau_fleurs_2023}
\BIBentryALTinterwordspacing
A.~Conneau, M.~Ma, S.~Khanuja \emph{et~al.}, ``{FLEURS}: {Few}-{Shot} {Learning} {Evaluation} of {Universal} {Representations} of {Speech},'' in \emph{2022 {IEEE} {Spoken} {Language} {Technology} {Workshop} ({SLT})}, Jan. 2023, pp. 798--805.
\BIBentrySTDinterwordspacing

\bibitem{kharitonov_textless-lib_2022}
\BIBentryALTinterwordspacing
E.~Kharitonov, J.~Copet, K.~Lakhotia \emph{et~al.}, ``textless-lib: a {Library} for {Textless} {Spoken} {Language} {Processing},'' in \emph{Proceedings of the 2022 {Conference} of the {North} {American} {Chapter} of the {ACL}: {Human} {Language} {Technologies}: {System} {Demonstrations}}.\hskip 1em plus 0.5em minus 0.4em\relax Hybrid: Seattle, Washington + Online: Association for Computational Linguistics, Jul. 2022, pp. 1--9.
\BIBentrySTDinterwordspacing

\bibitem{loshchilov_decoupled_2019}
\BIBentryALTinterwordspacing
I.~Loshchilov and F.~Hutter, ``\BIBforeignlanguage{en}{Decoupled {Weight} {Decay} {Regularization}},'' in \emph{\BIBforeignlanguage{en}{International {Conference} on {Learning} {Representations}}}, 2019.
\BIBentrySTDinterwordspacing

\bibitem{wolf_transformers_2020}
\BIBentryALTinterwordspacing
T.~Wolf, L.~Debut, V.~Sanh \emph{et~al.}, ``Transformers: {State}-of-the-{Art} {Natural} {Language} {Processing},'' in \emph{Proceedings of the 2020 {Conference} on {EMNLP}: {System} {Demonstrations}}.\hskip 1em plus 0.5em minus 0.4em\relax Online: Association for Computational Linguistics, Oct. 2020, pp. 38--45.
\BIBentrySTDinterwordspacing

\bibitem{hsu_liger_2025}
\BIBentryALTinterwordspacing
P.-L. Hsu, Y.~Dai, V.~Kothapalli \emph{et~al.}, ``Liger {Kernel}: {Efficient} {Triton} {Kernels} for {LLM} {Training},'' Jan. 2025, arXiv:2410.10989.
\BIBentrySTDinterwordspacing

\bibitem{papineni_bleu_2002}
\BIBentryALTinterwordspacing
K.~Papineni, S.~Roukos, T.~Ward \emph{et~al.}, ``Bleu: a {Method} for {Automatic} {Evaluation} of {Machine} {Translation},'' in \emph{Proceedings of the 40th {Annual} {Meeting} of the {ACL}}.\hskip 1em plus 0.5em minus 0.4em\relax Philadelphia, Pennsylvania, USA: Association for Computational Linguistics, Jul. 2002, pp. 311--318.
\BIBentrySTDinterwordspacing

\bibitem{rei_comet_2020}
\BIBentryALTinterwordspacing
R.~Rei, C.~Stewart, A.~C. Farinha \emph{et~al.}, ``{COMET}: {A} {Neural} {Framework} for {MT} {Evaluation},'' in \emph{Proceedings of the 2020 {Conference} on {EMNLP}}.\hskip 1em plus 0.5em minus 0.4em\relax Online: Association for Computational Linguistics, Nov. 2020, pp. 2685--2702.
\BIBentrySTDinterwordspacing

\end{thebibliography}

\end{document}